\documentclass[runningheads]{llncs}

\usepackage{eccv}

\usepackage{eccvabbrv}

\usepackage{graphicx}
\usepackage{booktabs}
\usepackage{multirow}
\usepackage{pifont}
\usepackage{wrapfig}

\usepackage[accsupp]{axessibility}  

\usepackage{hyperref}

\usepackage{orcidlink}

\newcommand{\EffiPerception}{EffiPerception}

\begin{document}

\title{\EffiPerception: an Efficient Framework for Various Perception Tasks} 

\titlerunning{\EffiPerception}

\author{Xinhao Xiang\inst{1}
\and
Simon Dr{\"a}ger\inst{1}
\and
Jiawei Zhang\inst{1}}

\authorrunning{X.Xiang et al.}

\institute{IFM Lab, Department of Computer Science, UC Davis CA 95616, USA
\email{xinhao@ifmlab.org   } 
\email{sdraeger@ucdavis.edu   }
\email{jiawei@ifmlab.org}}

\maketitle

\begin{abstract}
  The accuracy-speed-memory trade-off is always the priority to consider for several computer vision perception tasks. 
  Previous methods mainly focus on a single or small couple of these tasks, such as creating effective data augmentation, feature extractor, learning strategies, etc. These approaches, however, could be inherently task-specific: their proposed model's performance may depend on a specific perception task or a dataset.
  Targeting to explore common learning patterns and increasing the module robustness, we propose the {\EffiPerception} framework.
  It could achieve great accuracy-speed performance with relatively low memory cost under several perception tasks: 2D Object Detection, 3D Object Detection, 2D Instance Segmentation, and 3D Point Cloud Segmentation. 
  Overall, the framework consists of three parts: 
  (1) Efficient Feature Extractors, which extract the input features for each modality. (2) Efficient Layers, plug-in plug-out layers that further process the feature representation, aggregating core learned information while pruning noisy proposals. (3) The EffiOptim, an 8-bit optimizer to further cut down the computational cost and facilitate performance stability. 
  Extensive experiments on the KITTI, semantic-KITTI, and COCO datasets 
  revealed that {\EffiPerception} could show great accuracy-speed-memory overall performance increase within the four detection and segmentation tasks, in comparison to earlier, well-respected methods.
  
  \keywords{Computer Vision Perception Tasks \and Efficient Frameworks}
\end{abstract}

\section{Introduction}
\label{sec:intro}
\begin{table}[!htb]
    \centering
    \begin{tabular}{lcccc}
        \toprule
        \multirow{2}{*}{Models} & \multicolumn{2}{c}{2D perception} & \multicolumn{2}{c}{3D perception} \\
        \cmidrule(lr){2-3}\cmidrule(lr){4-5} & OD & IS & OD & PCS \\
        \midrule
        \cite{ViTAdapter, MobileNetV2, MobileNetV3, DeformableCNN, BiFormer, SwinTransformer, ConvXNet, PVTv2, MAE, SimultaneousODnSS, MultiTaskConsistencyForActiveLearning}                  & \ding{51} & \ding{51} & \ding{55} & \ding{55}\\
        \cite{SpatialTransformer, PVRCNN, 3DODnISfrom3DRangeAnd2DColorImages, LidarMultiNet, Joint3DISandODforAD, DASS}                       & \ding{55} & \ding{55} & \ding{51} & \ding{51}\\
        \cite{MultiTaskMultiSensorFusion}                      & \ding{51} & \ding{55} &	\ding{51}	& \ding{55} \\
        \cite{DiffBEV}                      & \ding{55} & \ding{51} &	\ding{51}	& \ding{55} \\
        \textbf{\EffiPerception}       & \ding{51} & \ding{51} & \ding{51} & \ding{51} \\
        \bottomrule
    \end{tabular}%
    \caption{EffiPerception is the first generic perception framework for the four perception tasks: 2D Object Detection (2D OD), 2D Instance Segmentation(IS), 3D Object Detection(3D OD), and Point Cloud Segmentation(PCS)}
    \label{tab:function-comparion}
\end{table}

Segmentation and object detection are two significant perception tasks in the fields of computer vision and robotics. 
With the emergence of deep neural networks\cite{Faster_RCNN,YOLO,SSD,Mask_RCNN,YOLOv7,SegmentAnything}, as well as large labeled image datasets\cite{ImageNet,COCO}, 2D object detection and instance segmentation systems from RGB images have been significantly improved. 
For applications related to robotics, such as self-driving cars, grasping, etc., a 2D object detection system is not adequate for their perception tasks. Thus, 3D object detection and segmentation systems have then been developed\cite{PointNet,VoxelNet,PVRCNN,TED,LoGoNet}.
Compared to the 2D tasks, 3D object detection and semantic segmentation need to collect additional depth information, which could be gained from the Lidar point cloud or other depth-only sensors.

Although some of these methods\cite{SparseRCNN,SegmentAnything,PVRCNN} could reach relatively high detection/segmentation accuracy, they may not take full consideration of inference speed and memory cost. 
Several methods are proposed to fill these gaps. For example, compact sparse R-CNN\cite{CompactSparseRCNN} speeds up the classic sparse R-CNN\cite{SparseRCNN} and reduces the memory cost by reducing iterative detection heads and simplifying the feature pyramid network. 
In 3D object detection, using sparse convolutions-based methods\cite{ConQueR,FSD++,FocalSConv} is also a common approach to speed it up.

Despite reaching relatively high performance, these methods mainly focus on a single or small couple of perception tasks. As a result, it could be inherently task/dataset-specific. For example, Sparse convolutions-based methods\cite{ConQueR,FSD++,FocalSConv} have shown high performance on 3D object detection, but it is still a question whether it could also reach the same competitive performance for 2D object detection and segmentation tasks. On the other hand, the compact strategy\cite{CompactSparseRCNN} well-used in sparse R-CNN may not provide the same performance gain in 3D perception tasks. 
\textit{Could there exist a unified framework that is efficient for various perception tasks, under corresponding input extractor and perception head?} 
To develop such a framework, three aspects could be re-investigated. 

The first aspect comes from the input feature extractors. 
For point cloud data, point-based, voxel-based, and range-based are the three main extraction approaches\cite{3DODinAD}. 
Before the extraction, some pre-processing techniques may also be applied, such as the random sampling\cite{StarNet}, set abstraction\cite{PointNet}, farthest point sampling(FPS)\cite{PointNet++}, segmentation guided filtering\cite{IPOD}, coordinate refinement\cite{Pointformer}, etc. 
Different extractors and pre-processing strategy choices could lead to huge differences. For example, increasing the number of context points will gain more representation power but at the cost of increasing memory consumption\cite{PointNet}. Suitable context radius in ball query is another key factor to balance the efficacy and efficiency of detection models\cite{PointNet++}. 
As for image data, besides using the raw RGB pixel data, extra depth information, stereo information, or multi-view input could also contribute to image extraction. 
Additional information increases the diversity of the input space though at the same time adds more challenges to the following feature learning. 
In this paper, we designed a unified extractor for each modality called the \textit{Efficient Feature Extractors}. 
For image feature extraction, we partition raw RGB images into smaller non-overlapping patches, flatten them, and conduct local learning.
For point cloud feature extraction, we adopt voxelization, random sampling, outlier removal, data augmentation, flattening them, and finally conducting local learning.
Under these settings, it could capitalize on the strengths of various previous approaches while easing their weaknesses, showing great performance increases with higher inference speed and lower training memory usage.

The second aspect arises from the following learning layers.
Previous approaches rely heavily on their task-driven or task-centered measures, such as adopting data augmentation\cite{Deepfusion,PointAugmenting,HSSDA}, or precisely defining its loss function combined with training strategies\cite{DETR,ConQueR}.
Take an example, DeepFusion\cite{Deepfusion} effectively aligns the transformed features from the input image and point cloud. One core technique named InverseAug is for inversing geometric-related augmentations to enable accurate geometric alignment between lidar points and image pixels. While this measure performs excellently on their focusing fusion-based 3D object detection tasks, it may not be generalized for other non-fusion tasks. 
Therefore, previous works usually adopted different operator and learning layers for camera-only, LiDAR-only, and camera-LiDAR fusion tasks. 
These tasks, however, have many similarities, suggesting a general unified framework should be a more efficient candidate. 
To this end, we propose the \textit{efficient layers}, which are more general layers that could act as the plug-in plug-out model in various perception tasks. 
Having extracted features from input extractors, the efficient layers further process them, aggregating core learned information while pruning noisy proposals. 
Being well-compatible with all four perception tasks, it further accelerates the inference speed while cutting training memory usage even more.

\begin{figure*}
    \includegraphics[width=\textwidth]{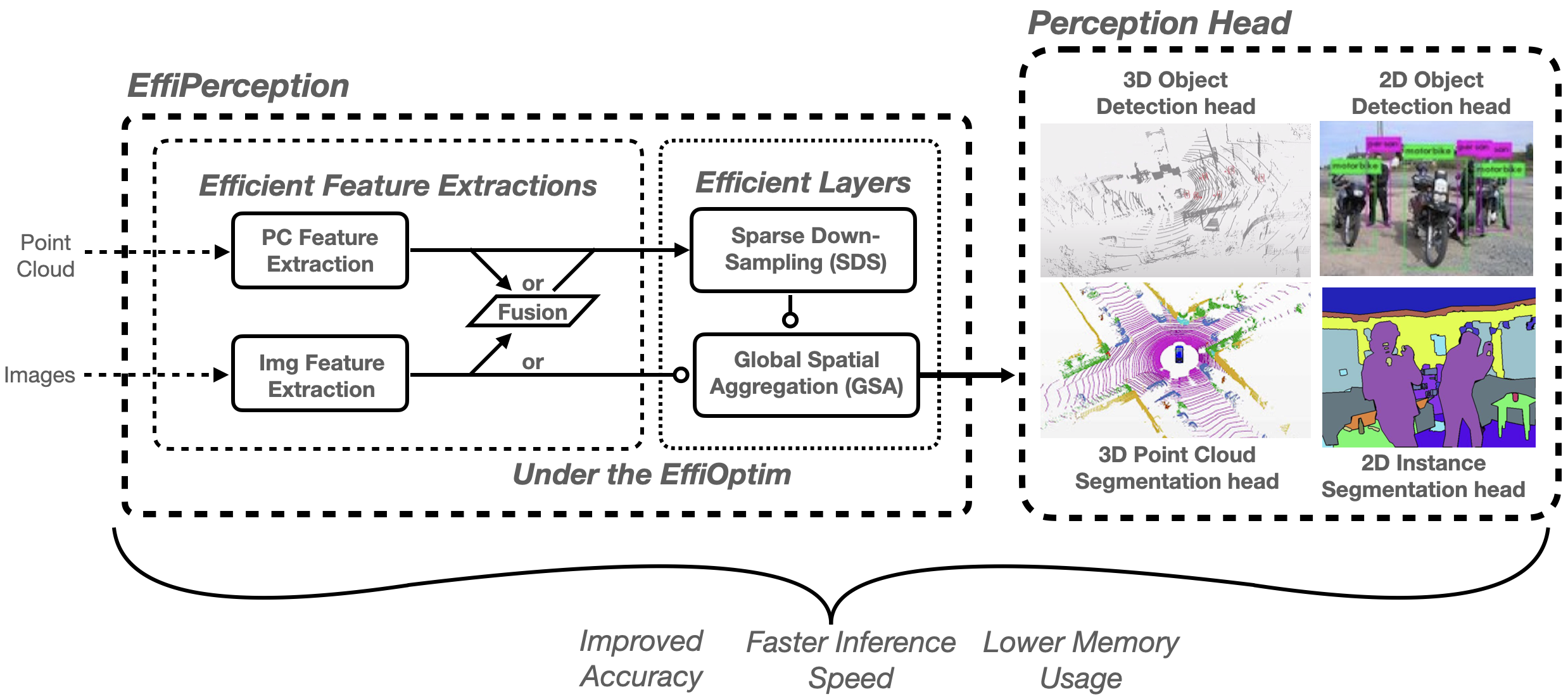}
    \caption{Overview of \EffiPerception, an efficient framework for various perception tasks}
    \label{fig:fig1}
\end{figure*}

The third aspect is the training framework, such as the optimizing scheme, training strategy, etc. A more efficient training framework would generate a more competitive output in a unified manner. 
By utilizing an 8-bit optimizer\cite{8BitOpti} that works on perception tasks, we named it \textit{the EffiOptim} to reduce training memory consumption while accelerating inference speed without compromising accuracy.
Its features of scalability and integration indicate great potential application scenes of training/deploying larger models on resource-constrained devices.



We conduct experience for various perception tasks under three datasets: COCO\cite{COCO} for 2D object detection and segmentation; KITTI\cite{Kitti} for 3D object detection; and Semantic-KITTI\cite{SemanticKITTI} for point cloud segmentation. By integrating various baselines with our \textit{EffiPerception} framework, its accuracy-speed performance could always be maintained at a high level, within relatively low memory cost.
This leads to a significant boost in the overall performance compared to their raw baselines. In addition, since we focus on improving the common parts among various tasks, the framework shows excellent robust scalability.
In summary, our main contributions are as follows:
\begin{itemize}

    \item We proposed an efficient framework called EffiPerception to achieve better accuracy-speed performance with relatively low memory cost under several perception tasks. 
    \item We investigated exploiting common layer patterns learning strategies among tasks, and proposed two generic layers: Sparse Down-Sampling and Global Spatial Aggregation, to increase the framework's robustness and efficiency.
    \item Extensive experiments are conducted on the COCO, KITTI, and semantic-KITTI datasets within four detection and segmentation tasks, showing fascinating accuracy-speed-memory overall performance increase. Its feature of scalability indicates potential application scenes of training/deploying larger models on resource-constrained devices. 
\end{itemize}

\section{Related Work}
\label{sec:related}

\subsection{Computer Vision Perception Tasks}

The goal of perception in Computer Vision is to understand the scene or feature in images of the real world\cite{PatternClsAndSceneAnalysis}. 
Preliminary feature learning including edge detection, pattern detection, and image classification, are the first few exploration\cite{LeNet, HOG, AlexNet}. 
In current main-steam perception tasks, \textit{object detection} models design detectors that post a yes/no question regarding the presence of an object class\cite{Faster_RCNN, YOLO, PVRCNN, TransFusion, LIFT, LoGoNet}. Another task is \textit{segmentation}, whose models extract the geometric description of an object by grouping pixels/point cloud\cite{Mask_RCNN, YOLACT, SegmentAnything, LESS, Cylinder3D, SphericalFormer}. 
These tasks cannot be done without various of sensors, such as cameras, LiDAR, RADAR, etc.  
Early models stick to 2D representations\cite{SSD, Faster_RCNN, Mask_RCNN}. With the popularization of 3D sensors like LiDAR as well as the development of GPU, many models could include depth information and present the 3D output\cite{PointNet++, VoxelNet, PVRCNN}. 
Today computer vision is well advanced to specific tasks, and could be well-used in several real application scenarios in robotics and self-driving cars\cite {Deepfusion, TransFusion, UniAD}. 
To broaden its usage into other harder application scenarios, the accuracy-speed-memory trade-off, as well as the versatility of these models that could reach high performance in various tasks, should be the following core issues to investigate.


\subsection{Efficiency of Deep Learning Models}

Typical technologies in efficient deep learning framework include network pruning\cite{learningWeiAndConnection, channelPruning, throughNetSlimming}, quantization\cite{DeepCompression}, efficient model architecture design\cite{MobileNets, ShuffleNetV2, EfficientNet, EfficientViT, SwinTransformer}, efficient optimizers\cite{8BitOpti, QLoRA, IntegralNN, PatrickStar}, and training techniques\cite{DistillingKnowledgeInNN, NetworkAugForTinyDL, TinyTL}. 
Besides manual designs, recent works also use AutoML techniques\cite{NASNet, EfficientNet} to automatically design\cite{OnceForAll}, prune\cite{AMC} and quantize\cite{APQ} neural networks.
The accuracy-speed-memory trade-off plays an important role in evaluating these models.

\subsection{Versatility of Deep Learning Models}

Previous approaches usually focus on a specific perception task\cite{Faster_RCNN, Mask_RCNN, PointNet++, PVRCNN}. 
They rely heavily on their task-driven or task-centered measures, such as designing a data augmentation strategy\cite{Deepfusion, ConQueR, PointAugmenting, HSSDA}, using different backbone learning layers for different tasks\cite{DETR, ConQueR, Pointformer}. 
Some methods could be well-adopted in two or three tasks, such as 2D object detection and instance segmentation\cite{MAE, PVTv2, SwinTransformer, ConvXNet}, 3D object detection and tracking\cite{CenterPoint, VoxelNet, VoxelNeXt}, 3D object detection and point cloud segmentation\cite{PVCNN, SphericalFormer, Robo3D},  
These measures, while performing excellently on their focusing tasks, may also not be generalized for others. As shown in \cref{tab:function-comparion}, we are the first versatile framework that could provide stable and extensive performance increase under all these four perception tasks.

\section{Proposed Method}
\label{sec:method}


\paragraph{Framework Overview}

\Cref{fig:fig1} shows the overall architecture of the proposed {\EffiPerception}. We focus on exploring the most common parts to increase the overall perception performance, to make it as general as possible. The {\EffiPerception} model consists of three core parts: the efficient feature extraction, the efficient layers, and the EffiOptim. It is a universal efficient framework for various perception tasks: 3D object detection, 3D point cloud segmentation, 2D object detection, and 2D instance segmentation. This framework accepts multi-modal inputs for 3D tasks, which include both RGB images and point clouds. For input images, they are defined as \(\mathcal{I} \in \mathbb{R}^{H_I \times W_I \times 3}\), where \(H_I\) and \(W_I\) denote the image height and width dimensions, respectively.
Meanwhile, the input point cloud is represented as a set of \(3 \mathrm{D}\) points \(\mathcal{P} \in \mathbb{R}^{H_P \times W_P \times D_P}\) in the \(H_P \cdot W_P \cdot D_P\) 3D space, where each point is a vector of its \((x, y, z)\)-coordinate.

Based on the specific perception task, this framework could use its corresponding feature transfer paths. For example, if the task is the Lidar-Only 3D object detection, then the input is only for the point cloud. In this way, only feature transfer paths relative to point cloud input, as well as the 3D object detection head, are activated. The possible optimal feature transfer paths are indicated by dotted lines. 

Not only for various input modalities and perception tasks, the framework is also compatible with a great number of existing methods. Using their same 3D or 2D feature extraction backbone (and neck if exists), as well as their perception head and loss function, we add the additional efficient layers and common feature extraction techniques under the efficient optimizer, EffiOpim. These relatively light modifications, however, could show excellent performance increase, no matter for inference speed, or the training memory usage. In the following part, we will describe the three core parts of {\EffiPerception} in detail. 


\subsection{Efficient Feature Extractors}
\begin{figure}
    \includegraphics[width=9.4cm]{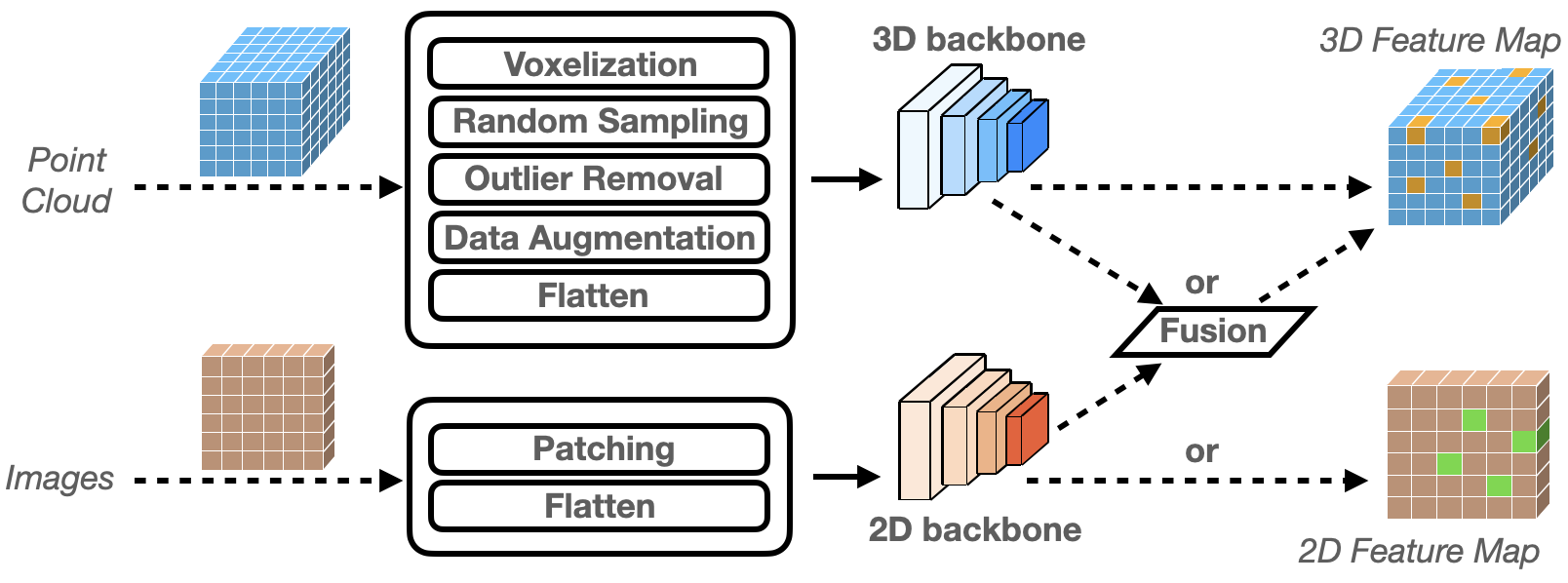}
    \centering
    \caption{The Efficient Feature Extraction framework}
    \label{fig:fig2}
\end{figure}
As shown in \Cref{fig:fig1}, given raw images \(\mathbf{I} \in \mathbb{R}^{H_I \times W_I \times 3}\) and/or 3D points \(\mathcal{P} \in \mathbb{R}^{H_P \times W_P \times D_P}\), we use a separate feature extractor to encode them.

\paragraph{Image Feature Extraction.}
For image features, we partition images into mini-patches before feeding them into the 2D backbone. The inherent dense feature of images often poses challenges related to memory consumption and computational efficiency. To mitigate these challenges, a promising approach involves partitioning the input image into smaller, non-overlapping patches before feeding it into the feature extractor\cite{ViT,SwinTransformer,FusionViT}. This strategy offers several advantages. Firstly, it allows for memory usage's significant reduction, as the computational and memory requirements scale with the input size. Secondly, patch partitioning enhances parallelism, enabling more efficient processing on modern hardware such as GPUs. Additionally, by decomposing the image into smaller, manageable regions, it facilitates the extraction of local context and intricate details. 

Formally, by setting each patch to have the side length $v_{cH}$ and $v_{cW}$ respectively, it partitions each image $\mathbf{I} \in \mathbb{R}^{H_I \times W_I \times 3}$ image into $N_c$ 2D patches $\mathbf{X_I} \in \mathbb{R}^{N_c \times\left(v_{cH}\cdot v_{cW} \cdot 3\right)}$, where $N_c=\frac{H_I}{v_{cH}}\frac{W_I}{v_{cW}}$ denotes the patch number.

Such partitioned image patches will be flattened into one-dimensional $v_{cH} \times v_{cW} \times 3$ features, then feed into a Multi-Layer Perceptron (MLP) to reach in total $N_c$ encoded features. For each encoded features $i$, we have
\begin{equation} 
\mathbf{x}_{c}^i = \operatorname{MLP}\left(\mathbf{x}_I^i\right).
\end{equation}
The MLP's parameters are shared by all patches so that they are encoded in the same way. Each encoded feature $\mathbf{x}_{c}^i$ has the same vector size $D_c$, which is also the output size of the MLP.

\paragraph{Point Cloud Feature Extraction.}
Compared to 2D features, 3D point cloud features contain an additional dimension, making learning more challenging. Inspired by\cite{VoxelNet,PVRCNN,3DifFusionDet}, voxelization is adopted before feeding into the 3D backbone. The inherent sparsity of point clouds brings lots of useless computation and resource occupation. Voxelization transforms the discrete and scattered point cloud data into a structured grid representation, where each voxel represents a fixed 3D volume within the space. It could significantly reduce memory usage and enhance parallelism. Furthermore, by aggregating points within each voxel, it promotes local context understanding, facilitating feature extraction processes.

To do it, we divide the raw point cloud in the whole 3D space into little cubic, each with side length $P_l$. Those empty cubic are removed to cut down the computation burden. Setting each cubic has the side length $v_{lH}$, $v_{lW}$ and $v_{lD}$ respectivel, it transforms the whole $\mathcal{P} \in \mathbb{R}^{H_P \times W_P \times D_P}$ point cloud space into $N_l$ 3D cubics $\mathbf{X_P} \in \mathbb{R}^{N_l \times\left(v_{lH}\cdot v_{lW} \cdot v_{lD}\right)}$, where: $N_l=\frac{H_P}{v_{lH}}\frac{W_P}{v_{lW}}\frac{D_P}{v_{lD}}$ denotes the cubic number.
Furthermore, we conduct random sampling for those cubic having large point cloud numbers, to a fixed smaller number, T. This sampling strategy is not only for computational savings but to decrease the imbalance of points between the cubic, which reduces the sampling bias. For $\mathbf{X_P}^\prime=\left\{\mathbf{x_P^i}=\left[x_i, y_i, z_i\right]^T \in \mathbb{R}^3\right\}_{i=1 \ldots t}$ as a non-empty cubic containing $t \leq T$ LiDAR points, where $\mathbf{x_P^i}$ contains XYZ coordinates for the $i$-th point. We also compute the local mean as the centroid of all the points in $\mathbf{X_P}^\prime$, denoted as $\left(c_x, c_y, c_z\right)$. Then we augment each point $\mathbf{x_P^i}$ with the relative offset w.r.t. the centroid to get input feature set $\mathbf{X_{Pin}}=\left\{\hat{\mathbf{x_P^i}}=\left[x_i, y_i, z_i, x_i-c_x, y_i-c_y, z_i-c_z\right]^T \in \mathbb{R}^6\right\}_{i=1 \ldots t}$.
Each cubic will then be flattened into one-dimensional $v_{lH} \times v_{lW} \times v_{lD}$ features, and feed into an MLP. For each encoded features $i$, we have: 
\begin{equation} 
\mathbf{x}_{l}^{i^{\prime}} = \operatorname{MLP}(\hat{\mathbf{x_{P}^i}}).
\end{equation}

After voxelization and/or patching, learned features will be transferred into the corresponding backbones. Their output feature maps will be further utilized in the efficient layers. If fusion exists, its fused feature will be treated as 3D feature map. Note that the operation similarity makes the fusion more smooth.

\subsection{Efficient Layers}
\begin{figure}
    \includegraphics[width=12cm]{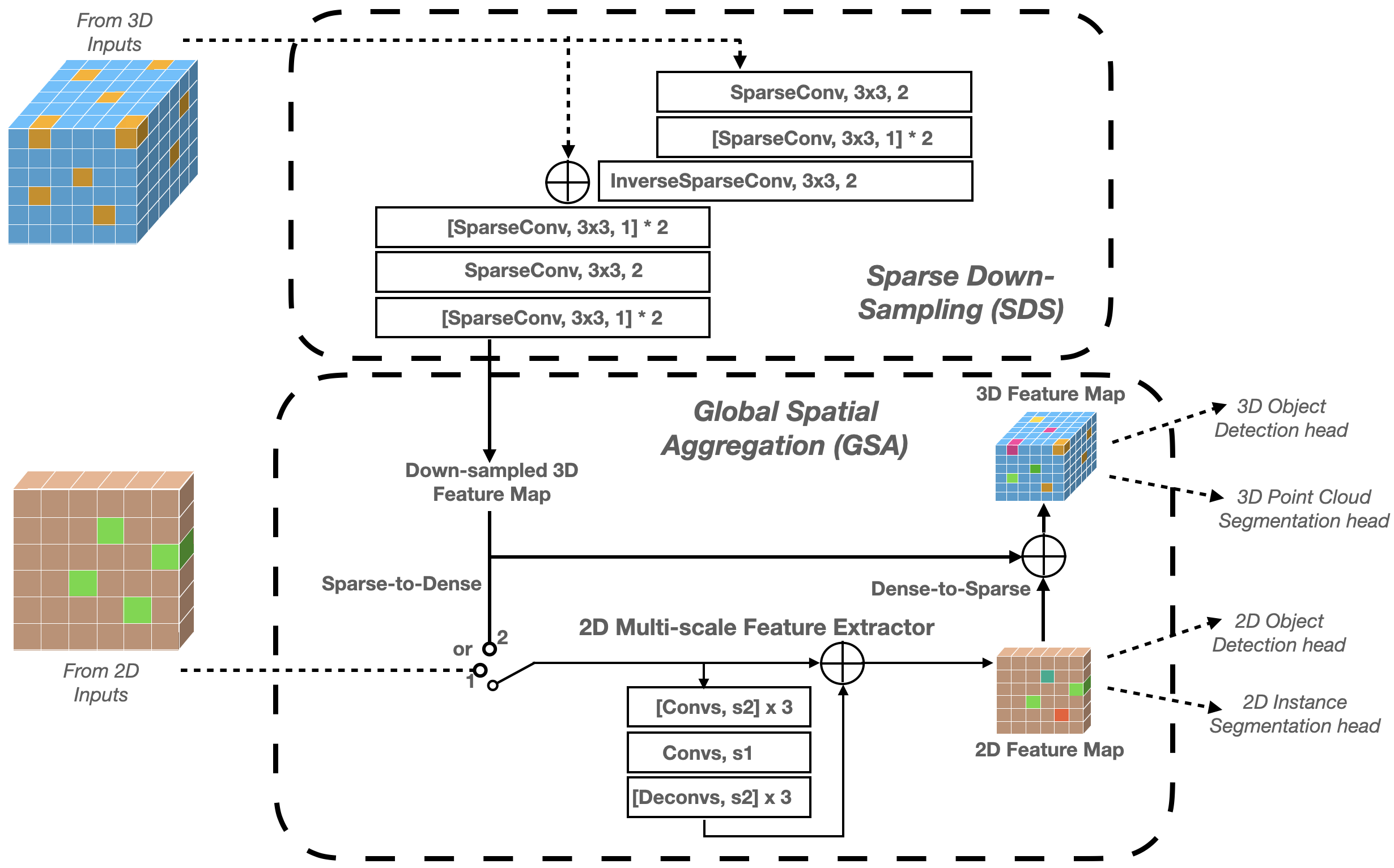}
    \centering
    \caption{The Efficient Layers}
    \label{fig:fig3}
\end{figure}
Efficient Layers are plug-in plug-out layers that further process the feature representation, aggregating core learned information while pruning noisy proposals. We explore the most common approaches that could be well-compatible for all four perception tasks. 

We first analyze the core difference between 2D and 3D perception tasks. In 2D image space, occlusion, spatial layout ambiguity, and Scale\cite{ReliableScaleEstimation} are three main problems\cite{Joint3DISandOD,DLinCV}. We do need some operations to aggregate spatial information. 
In 3D point cloud space, while it is also necessary to accumulate valid proposals, object features could become hugely separable. Directly adopting the aggregation from the point cloud not only triggers dimension mismatch issues but also produces unsatisfied results. In addition, raw 3D feature representations are much more complex than 2D's, so a well-designed intermediate procedure is urgently required.
Therefore, we designed Sparse Down-sampling layers for 3D input feature map. These layers help preliminary aggregate and down-sample raw 3D feature representations from the backbone. These processed features then feed into the Global Spatial Aggregation(GSA) layers. In these layers, features from 3D and 2D with be smoothly aggregated. The following will introduce the two groups of layers one by one. 


\paragraph{Sparse Down-Sampling for 3D features.}

Shown in the upper part in \Cref{fig:fig3}, these layers are composed of several sparse convolution and Inverse sparse convolution layers. Since firstly introduced\cite{SSC} and applied in 3D perception tasks\cite{SECOND}, many previous methods have shown that sparse convolution operations could speed up traditional convolutional operations while using smaller memory cost\cite{SWFormer,RT3DOD,SST,FSD++}. Our design follows the down-sampling, aggregation, and down-sampling routing. 3D input features will first do the down-sampling by getting through the sparse convolutional layer with stride 2. After two more sparse convolutional layers with stride 1, an inverse sparse convolutional layer comes to restore its feature dimension, which could be aggregated with the raw input feature. One further sparse convolutional layer with stride 1 and another downsampling recipe follow the concatenation. In this way, 3D feature space could smoothly down-sample its space while keeping more dense-like important information. 

Such a down-sampled 3D feature map could be more inherent-nature consistent with the 2D feature space, especially after conducting the sparse to dense operation in the Global Spatial Aggregation (GSA), which will be introduced in the next part. 
It is worth indicating that the Sparse Down-sampling is also important even if there is no need to combine it with the 2D features. Under pure-3D perception tasks, the Sparse Down-sampling operation follows the nature of 3D features, filtering in more valuable features, making them more aggregated, and gradually reducing the space size.

\paragraph{Global Spatial Aggregation.}

Besides using sparse convolutions, another common approach to handling 3D perception tasks is to first compress the 3D feature map into a 2D feature map, either as Bird-Eye-View(BEV)\cite{PointPillars,BEVFusion,TransFusion,TransFuser} or Range View(RV)\cite{MV3D,RSN,LaserNet}, then to use a 2D backbone or neck to extract the proposals. For example, PointPillar\cite{PointPillars} adopts a top-down backbone with downsampling, upsampling, and concatenation operations to exploit its feature mining. Inspired by these design\cite{LidarMultiNet}, we develop Global Spatial Aggregation (GSA). It could extract global contextual features from 3D sparse tensors by projecting it to $2 \mathrm{D}$ dense $\mathrm{BEV}$ feature map and then converting the dense feature map back after applying 2D multi-scale feature extractor.

As illustrated in the lower part of \Cref{fig:fig3}, given the 3D sparse tensor output by the encoder, we first project it into a dense $\mathrm{BEV}$ feature map $\mathcal{H}^{\text {sparse }} \in \mathbb{R}^{C \times M^{\prime}} \rightarrow \mathcal{H}^{\text {dense }} \in \mathbb{R}^{C \times \frac{D}{d z} \times \frac{H}{d x} \times \frac{W}{d y}}$, where $d$ is the down-sampling ratio and $M^{\prime}$ is the number of valid voxels in the last scale. After that, we concatenate the features across the height dimension together to form a $2 \mathrm{D} B E V$ feature map $\mathcal{H}^{\text {bev }} \in \mathbb{R}^{\left(C * \frac{D}{d_z}\right) \times \frac{H}{d_x} \times \frac{W}{d_y}}$. Then, we use a 2D multi-scale CNN to further extract contextual information. Lastly, we reshape the encoded BEV feature representation to the dense voxel map, then transform it back to the sparse voxel feature following the reverse dense-to-sparse conversion. To make the module be more robust, the transformed sparse voxel feature will be concatenated with the sparse feature before the sparse-to-dense operation.

Benefiting from GSA, our architecture could significantly enlarge the receptive field, which plays an important role in semantic segmentation. These operations not only are greatly effective for extracting 2D features, but are also natural-friendly for the compressed 3D features. In addition, both 2D tasks and 3D tasks could gain well-structured information from each other.

The features from generated 2D and 3D feature maps will be finally transferred into the corresponding perception head, according to the task. The 3D object detection and 3D point cloud segmentation head are followed from the 3D feature map, while from the 2D feature map, 2D instance segmentation and 2D object detection head are performed.

\subsection{The EffiOptim}

Besides these efficient layers and feature extractors, we also adopt an efficient optimizer, named EffiOptim, to further cut down training memory usage as well as speed up inference speed without sacrificing performance.

The development of efficient optimization techniques\cite{QLoRA,IntegralNN} for deep learning models has been a critical area of research, particularly in the context of memory usage and computational speed. In this regard, the emergence of 8-bit optimizers\cite{8BitOpti} has garnered significant attention due to their potential to reduce memory footprint and accelerate training and inference processes while maintaining performance levels comparable to traditional 32-bit optimizers. The EffiOptim is a 8-bit optimizers that work on perception tasks. This approach has the potential to further improve the efficiency of the {\EffiPerception} framework and enable us to train and deploy larger models on resource-constrained devices.




\begin{table*}[htbp]
    \centering
    \begin{tabular}{lccccccccc}
        \toprule
        \multirow{2}{*}{Model} & \multicolumn{3}{c}{$\text{mAP}_{\text{BEV}}(\text{IoU} = 0.7)$} & \multicolumn{3}{c}{$\text{mAP}_{\text{3D}}(\mathrm{IoU}=0.7)$} & Memory & Runtime\\
        \cmidrule(lr){2-4}\cmidrule(lr){5-7} & Easy & Med & Hard & Easy & Med & Hard & $(\mathrm{M})$ & $(\mathrm{ms})$\\
        \midrule
        PV-RCNN\cite{PVRCNN}  & 86.2 & 84.8 & 78.7 &	N/A	& N/A  & N/A & 16046 & 59.2 \\
        PointRCNN\cite{PointRCNN}                     & 87.9 & \underline{90.2} & 85.5 & 88.6 & 88.1 & 77.4 & 13289 & 100.1 \\
        CT3D\cite{ct3d}                          & 88.5 & 86.1 & 79.0 & 90.5 & 87.1 & 79.0 & N/A & N/A \\
        3D-SSD\cite{3DSSD}                        & 89.7 & 89.5 & 78.7 & 89.4 & 87.5 & 78.4 &  \textbf{6032} & \textbf{38.9} \\
        Part-$A^2$-free\cite{PartA2Point}                & 88.0 & \underline{90.2} & \textbf{85.9} & 89.0 & \textbf{88.5} & 78.4 & 12839 & 80.8 \\
        TED\cite{TED}                           & \underline{91.6} & 85.3 & 80.7 & \underline{91.5} & 85.7 & \underline{82.2} & 8820 & 105.5 \\
        LoGoNet\cite{LoGoNet}                       & \textbf{91.8} & 85.0 & 80.7 & \textbf{91.8} & 85.0 & \textbf{82.4} & 15023 & 96.3 \\
        \textbf{\EffiPerception}       & 89.9 & \textbf{90.5} & \underline{85.6}  & \underline{91.5} & \underline{88.2} & \textbf{82.4}  & \underline{7293} & \underline{42.0} \\
        \bottomrule
    \end{tabular}%
    \caption{Comparison of attained validation mAP (in \%) on KITTI with $\text{IoU} = 0.7$.}
    \label{tab:results-kitti}
\end{table*}

\begin{table*}[htbp]
    \centering
    \begin{tabular}{l|c|p{10pt}p{10pt}p{10pt}p{10pt}p{10pt}p{10pt}p{10pt}p{10pt}p{10pt}p{10pt}p{10pt}p{10pt}p{10pt}p{10pt}p{10pt}p{10pt}p{10pt}p{10pt}p{10pt}}
        \toprule
        Method & mIoU & \rotatebox{90}{car} & \rotatebox{90}{bicycle} & \rotatebox{90}{motorcycle} & \rotatebox{90}{truck} & \rotatebox{90}{other-vehicle} & \rotatebox{90}{person} & \rotatebox{90}{bicyclist} & \rotatebox{90}{motorcyclist} & \rotatebox{90}{road} & \rotatebox{90}{parking} & \rotatebox{90}{sidewalk} & \rotatebox{90}{other-ground} & \rotatebox{90}{building} & \rotatebox{90}{fence} & \rotatebox{90}{vegetation} & \rotatebox{90}{trunk} & \rotatebox{90}{terrain} & \rotatebox{90}{pole} & \rotatebox{90}{traffic-sign}\\
        \midrule
        PointNet++\cite{PointNet++} & 20.7 & 72 & 42 & 19 & 6 & 62 & 54 & 1 & 2 & 0 & 0 & 47 & 14 & 30 & 1 & 1 & 0 & 17 & 6 & 9 \\
        SqueezeSegV3\cite{SqueezeSegV3} & 56.4 & \textbf{92} & 75 & 63 & 26 & 89 & 93 & 30 & 39 & 37 & 33 & 82 & 59 & 65 & 46 & 46 & 20 & 59 & 50 & 59 \\
        Cylinder3D\cite{Cylinder3D} & 69.2 & \textbf{92} & 77 & 65 & 32 & 91 & 97 & 51 & 68 & 64 & 59 & 86 & 73 & 70 & 74 & 69 & 48 & 67 & 62 & 66 \\
        SPVNAS\cite{SPVNAS} & 67.8 & 90 & 75 & 68 & 22 & 92 & 97 & 57 & 51 & 50 & 58 & 86 & 73 & 71 & 67 & 67 & 50 & 67 & 64 & 67 \\
        PVKD\cite{PVKD} & 71.3 & \textbf{92} & 71 & \textbf{78} & \underline{43} & 92 & 97 & \textbf{68} & \underline{69} & 54 & 60 & \underline{87} & 74 & \textbf{72} & 75 & 74 & 51 & 69 & \underline{67} & 66 \\
        SphericalFormer\cite{SphericalFormer} & 75.0 & \textbf{92} & \underline{78} & 70 & 41 & \textbf{94} & \textbf{98} & \underline{60} & \textbf{70} & \underline{71} & \textbf{68} & \underline{87} & \textbf{75} & \textbf{72} & \underline{79} & \textbf{80} & \textbf{75} & \textbf{73} & 65 & \underline{73} \\
        
        \textbf{\EffiPerception} & \textbf{75.8} & 91 & \textbf{85} & \underline{71} & \textbf{48} & \underline{93} & \textbf{98} & 58 & \underline{69} & \textbf{73} & \underline{66} & \textbf{88} & \textbf{75} & 69 & \textbf{83} & \underline{78} & \underline{70} & \underline{71} & \textbf{68} & \textbf{75}\\
        \bottomrule
    \end{tabular}%
    \caption{Comparison of IoU (in \%) on Semantic KITTI \textit{test} set}
    \label{tab:results-semantic-kitti}
\end{table*}

\begin{table*}[htbp]
    \centering
    \begin{tabular}{lccccccccc}
        \toprule
        Model & box $\mathrm{AP}$ & $\mathrm{AP}_{50}$ & $\mathrm{AP}_{75}$ & $\mathrm{AP}_s$ & $\mathrm{AP}_m$ & $\mathrm{AP}_l$ & Memory$(\mathrm{M})$ & Runtime$(\mathrm{ms})$ \\
        \midrule
        MobileNetV2\cite{MobileNetV2} & 28.3 & 46.7 & 29.3 & 14.8 & 30.7 & 38.1 & 11283 & 62.3 \\
        MobileNetV3\cite{MobileNetV3} & 29.9 & 49.3 & 30.8 & 14.9 & 33.3 & 41.1 & \underline{10028} & 54.2 \\
        SPOS\cite{SPOS} & 30.7 & 49.8 & 32.2 & 15.4 & 33.9 & 41.6 & 13827 & 102.9 \\  
        MNASNet-A2\cite{MNASNet} & 30.5 & 50.2 & 32.0 & 16.6 & 34.1 & 41.1 & 13726 & 128.1 \\
        FairNAS-C\cite{FairNAS}  & 31.2 & 50.8 & 32.7 & 16.3 & 34.4 & 42.3 & 22192 & 101.3 \\
        MixNet-M\cite{MixNet}  & 31.3 & 51.7 & 32.4 & 17.0 & 35.0 & 41.9 & 18279 & 77.2 \\
        EfficientViT-M4\cite{EfficientViT} & 32.7& 52.2& 34.1 & 17.6 & 35.3 & 46.0 & 11938 & 52.3 \\
        EfficientDet-D3\cite{EfficientDet} & 46.8 & 65.9 & 51.2 & N/A & N/A & N/A & 13827 & \underline{40.2} \\
        ViT-Adapter-B\cite{ViTAdapter} & 47.0 & 68.2 & 51.4 & N/A & N/A & N/A & 16927 & 76.7 \\
        YOLOv7\cite{YOLOv7} & \underline{51.2} & \textbf{69.7} & \textbf{55.5} & \underline{35.2} & \underline{56.0} & \textbf{66.7} &  12023  & 45.4 \\
        \textbf{\EffiPerception} & \textbf{52.3} & \underline{69.3} & \underline{54.6} & \textbf{35.3} & \textbf{57.3} & \textbf{64.0} & \textbf{9946} & \textbf{36.2}\\
        \bottomrule
    \end{tabular}%
    \caption{Comparison of attained validation AP (in \%) on COCO \texttt{val2017}}
    \label{tab:results-coco-od}
\end{table*}

\begin{table*}[htbp]
    \centering
    \begin{tabular}{lcccccccc}
        \toprule
        Model & mask $\mathrm{AP}$ & $\mathrm{AP}_{50}$ & $\mathrm{AP}_{75}$ & $\mathrm{AP}_s$ & $\mathrm{AP}_m$ & $\mathrm{AP}_l$ & Memory$(\mathrm{M})$ & Runtime$(\mathrm{ms})$ \\
        \midrule
        FCIS\cite{FCIS}  & 29.5 & 51.5 & 30.2 & 8.0 & 31.0 & 49.7 & 14192 & 151.5 \\
        YOLACT-550\cite{YOLACT}  & 29.8 & 48.5 & 31.2 & 9.9 & 31.3 & 47.7 & 14872 & \textbf{42.1} \\
        RetinaMask\cite{RetinaMask} & 34.7 & 55.4 & 36.9 & 14.3 & 36.7 & 50.5 & 19028 & 166.7 \\
        Mask R-CNN\cite{Mask_RCNN} & 35.7 & 58.0 & 37.8 & 15.5 & 38.1 & 52.4 & 20397 & 116.3 \\
        PA-Net\cite{PANet} & 36.6 & 58.0 & 39.3 & 16.3 & 38.1 & 53.1 & 18239 & 112.8 \\
        MS R-CNN\cite{MaskScoringRCNN}  & 38.3 & 58.8 & 41.5 & 17.8 & 40.4 & 54.4 & 17029 & 86.3 \\
        BoxInst\cite{BoxInst} & 35.0 & 59.3 & 35.6 & 17.1 & 37.2 & 48.9 & 19827 & 93.2\\
        SOLOv2\cite{SOLOv2} & 39.7 & 60.7 & 42.9 & 17.3 & 42.9 & 57.4 & \underline{13827} & 63.2\\
        ViT-Adapter-B\cite{ViTAdapter} & \underline{41.8} & \textbf{65.1} & \underline{44.9} & N/A & N/A & N/A & 15729 & 127.4\\ 
        \textbf{\EffiPerception} & \textbf{42.1} & \underline{64.6} & \textbf{45.8} & \textbf{20.2} & \textbf{45.2} & \textbf{59.6} & \textbf{10382} & \underline{53.8}\\
        \bottomrule
    \end{tabular}%
    \caption{Comparison of attained validation AP (in \%) on COCO \texttt{val2017}}
    \label{tab:results-coco-is}
\end{table*}

\section{Experiments}
\label{sec:exp}

\subsection{Experimental Setup}

\paragraph{Datasets.}

To demonstrate the effectiveness of {\EffiPerception}, we present results on the KITTI~\cite{Kitti}, Semantic KITTI\cite{SemanticKITTI}, and COCO\cite{COCO}. 

KITTI is a 3D object detection benchmark. It is divided into 7,481 training samples and 7,518 testing samples. The training samples are commonly divided into a training set (3,712 samples) and a validation set (3,769 samples) following~\cite{KITTIdataSplit}, which we adopt. Before being fed into the models, the dataset is augmented by random 3D flips, random rotation, scale, and translation, and shuffled the point data. We use it to conduct 3D Object Detection experiments.
Semantic KITTI is a point-level re-annotation of the LiDAR part of the KITTI dataset. It has a total of 43551 scans sampled from 22 sequences collected in different cities in Germany. It has 104452 points per scan on average and each scan is collected by a single Velodyne HDL-64E laser scanner. We follow SemanticKITTIs subset split protocol and use ten sequences for training, one for validation, and the rest of them for testing. Our tests with 3D point cloud segmentation are carried out with it.
COCO 2017 is a 2D image dataset with 118K training images. We carry out 2D object detection and instance segmentation using it.



\paragraph{Implementation Details.}

We implement {\EffiPerception} using the MMDetection3D library~\cite{MMDet3D} and Detectron2\cite{Detectron2}.
The model is optimized using the Adam~\cite{Adam} optimizer with a learning rate of 0.0001, optimizer momentum $(\beta_1, \beta_2) = (0.9, 0.999)$, and a dropout rate of 0.3.
We train the model on an NVIDIA RTX A6000 GPU for 50 epochs and validate after each epoch.
On KITTI, we use $[2,46.8] \times[-30.08,30.08] \times[-3,1]$ for the range and $[0.16,0.16,0.16]$ for the voxel size for the $x, y$, and $z$ axes, respectively. On Semanitic KITTI, we use $[-51.2,51.2] \times[-51.2,51.2] \times[-4, 2.4]$ for the range and $0.05$ for the voxel size. The window size is set to $\left[120 m, 2^{\circ}, 2^{\circ}\right]$ for $(r, \theta, \phi)$.
For any possible direct set prediction, we set the number of proposal boxes to 300.

\subsection{Comparison with State-of-the-art models}


We conduct experiments for 3D object detection, 3D point cloud segmentation, 2D object detection, and 2D instance segmentation. 
\cref{tab:results-kitti,tab:results-semantic-kitti,tab:results-coco-od,tab:results-coco-is} shows their results correspondingly. In the 3D object detection task, we conduct experiments on the KITTI 3D object detection benchmark. 
All models are trained with batch size of 2, and their training memory and inference running time are also reported.
Our model is conducted with the SST\cite{SST} model.
Following the standard KITTI evaluation protocol ($\text{IoU} = 0.7$) for measuring the detection performance, it shows the mean average precision (mAP) scores for the {\EffiPerception} method compared to the state-of-the-art methods on the KITTI validation set using 3D and bird’s eye view (BEV) evaluation. 
We bold-face the two best-performing models for each task.
As indicated in \cref{tab:results-kitti}, our approach shows significant overall performance improvements compared to the baselines, not only on the accuracy but also for the training memory usage and inference runtime. 

The 3D point cloud segmentation experiments are conducted in the semantic-KITTI benchmark whose \textit{test} set results are shown in \cref{tab:results-semantic-kitti}. All models are trained with batch size of 2. 
The LaserMix\cite{LaserMix} model is applied to perform our model.
Consistently outperforms others by a large margin, our method yields 75.8\% mIoU, a new state-of-the-art result. Thanks to the capability of directly aggregating long-range information, our method significantly outperforms the models based on classic sparse convolution\cite{Cylinder3D, SPVNAS}.

Besides, \cref{tab:results-coco-od,tab:results-coco-is} shows the results of 2D object detection and instance segmentation experiments conducted on the COCO dataset. All models are trained with 16 batch size and their training memory and inference running time are also reported. 
Our model adopts Sparse-RCNN\cite{SparseRCNN} and EfficientViT\cite{EfficientViT} correspondingly as the backbone of these two tasks.  
In the detection task, EffiPerception exhibits better performance in accuracy (52.3 AP) and shorter inference time (36.2ms), with shorter memory usage(9946M). 
For the segmentation task, EffiPerception also outperforms well-established detectors, such as Mask-RCNN\cite{Mask_RCNN} and YOLACT\cite{YOLACT}, by a large margin. Its shorter inference time and lower memory usage reveal a wide range of potential usages.



\begin{figure}
    \includegraphics[width=12cm]{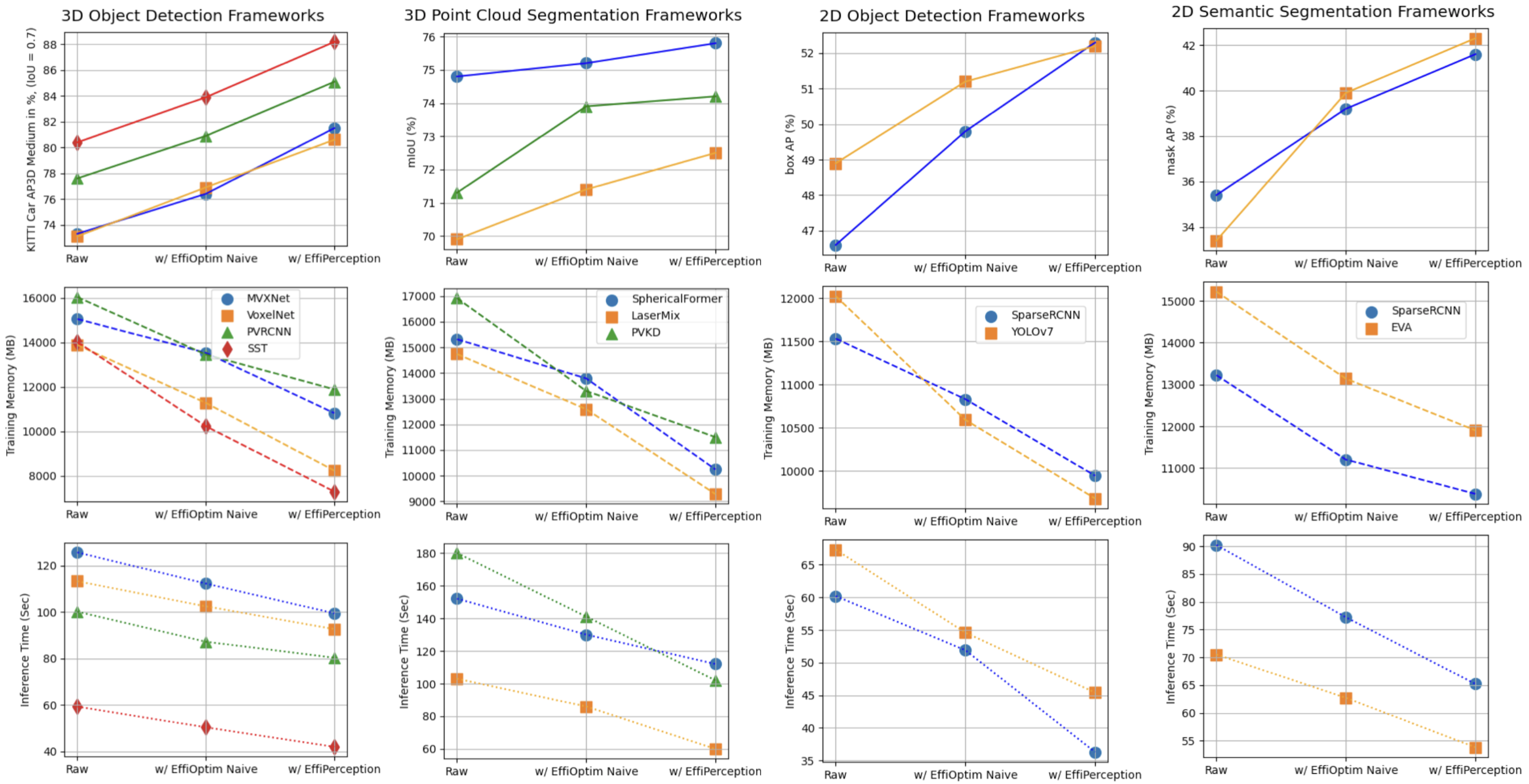}
    \centering
    \caption{Experiment Results}
    \label{fig:fig6}
\end{figure}

\subsection{EffiPerception is a generic perception method}
\label{sec:generic-model}


\begin{wraptable}{r}{0pt} 
    \begin{tabular}{ccccc}
        \toprule
        \multirow{2}{*}{ SDS } & \multirow{2}{*}{ GSA } &
        \multicolumn{3}{c}{ $\text{mAP}_{\text{BEV}} (\text{IoU}=0.7)$ } \\ 
        \cmidrule(lr){3-5}
          &  & Easy & Med & Hard \\
        \midrule
        \ding{55} & \ding{55} & 88.5 & 85.2 & 78.4\\
        \ding{51} & \ding{55} & 89.4 & 86.5 & 80.5\\
        \ding{55} & \ding{51} & 89.8 & 87.2 & 80.3\\
        \ding{51} & \ding{51} & $\mathbf{91.5}$ & $\mathbf{88.2}$ & $\mathbf{82.4}$ \\
        \bottomrule
    \end{tabular}
    \caption{Ablation studies on Sparse Down-Sampling (SDS) and Global Spatial Aggregation (GSA). 
    }
    \label{tab:ablation-components}
\end{wraptable}

Next, we examine how generic our method is by plugging it into prevalent perception frameworks. 
Among these four perception tasks, we conduct \textit{eleven} pairs of comparisons in total, each of which is between a single-modal method and its EffiPerception-equipped counterpart. 
To further clearly show the performance gain from the neural network model, we add a middle column to report the modal performance combined only with the EffiOptim (as `w/EffiOptim native'). 

Those ten models' performances are shown in \cref{fig:fig6}, where each column figure group corresponds to one task. 
Owning higher accuracy, smaller memory usage, and lower inference time, EffiPerception indicates a consistent improvement for all single-modal baselines among all the four down-steaming tasks. 
These results indicate EffiPerception is generic and can be potentially applied to various perception frameworks.

\subsection{Impact of Sparse Down-Sampling and Global Spatial Aggregation}
\label{sec:ablation-studies-1}

In this section, we show the effectiveness of both proposed components, Sparse Down-Sampling and Global Spatial Aggregation. 
\cref{tab:ablation-components} shows the experiment result of EffiPerception in 3D object detection task. 
We observe that both components can improve the performance over the single-modal baseline. 
In particular, the gain by Global Spatial Aggregation is more prominent. 
For example, without Global Spatial Aggregation, the performance of $mAP_{BEV}$ drops drastically from 88.2 to 86.5 in the Medium level.

\subsection{EffiPerception is more robust}
\label{sec:ablation-studies-2}

\begin{wraptable}{r}{0pt} 
    \begin{tabular}{lcccl}
        \toprule 
        \multirow{2}{*}{ Corruptions } & \multirow{2}{*}{ Modal } & 
        \multicolumn{3}{c}{ $\text{mAP}_{\text{BEV}} (\text{IoU}=0.7)$ } \\ 
        \cmidrule(lr){3-5}
          &  & Easy & Med & Hard \\
        \midrule 
        No Corruption & TED\cite{TED} & 91.5 & 85.7 & 82.2 \\
        Laser Noise & TED & 87.2 & 83.6 & 77.8\textit{(-4.4)} \\
        \midrule 
        No Corruption & LoGoNet\cite{LoGoNet} & 91.8 & 85.0 & 82.4 \\
        Laser Noise & LoGoNet & 87.4 & 83.5 & 78.6 \textit{(-3.8)} \\
        \midrule 
        No Corruption & EffiPerception & 91.5 & 88.2 & 82.4 \\
        Laser Noise & EffiPerception & 90.5 & 87.8 & 81.6\textit{(-0.8)} \\
        Pixel Noise & EffiPerception & 90.4 & 87.5 & 80.9\textit{(-1.5)} \\
        Pixel+Laser Noise & EffiPerception & 89.4 & 86.8 & 80.0\textit{(-2.4)} \\
        \bottomrule
    \end{tabular}
    \caption{Model robustness against input corruptions.}
    \label{tab:ablation-robust}
\end{wraptable}

Robustness is a pivotal metric as it ensures that the model can maintain its performance and provide reliable predictions even when faced with noisy, incomplete, or adversarial data, which is crucial for applications where safety, security, and accuracy are of paramount importance. 

In this subsection, we examine the model robustness against corrupted input\cite{CorruptionsBenchmarking}.
We test the model robustness on the validation set against common corruptions, Laser Noise (randomly adding noise to lidar reflections), and Pixel Noise (randomly adding noise to camera pixels). 
As shown in \cref{tab:ablation-robust}, with the presence of corruptions, our EffiPerception models, in general, are much more robust than their counterparts. The Lase or Pixel Noise corruption can hardly drag down the performance of our EffiPerception method (with only 0.8 / 1.5 $\text{mAP}_{\text{BEV}}$ Hard drop). 
Even when applying both Laser and Pixel Noise corruptions, the performance drop is still marginal (2.4 $\text{mAP}_{\text{BEV}}$ Hard drop). Meanwhile, the comparing LoGoNet and TED model drops around 4 APH $\text{mAP}_{\text{BEV}}$ Hard simply applying the Laser Noise corruption.



\section{Conclusion}
\label{sec:conclusion}
This paper presents EffiPerception, an efficient framework for various perception tasks. It focuses on learning common patterns to improve the overall performance under the feature extractor level, learning layers level, and optimizer level. Compared to well-established models, EffiPerception achieves great accuracy-speed performance with relatively low memory cost under several perception tasks: 2D Object Detection, 3D Object Detection, 2D Instance Segmentation, and 3D Point Cloud Segmentation.  
Its generic adaptive nature, extensive task-applicable features, as well as robust learning results make it promising for potential future usage.

%
%
\bibliographystyle{splncs04}
\bibliography{main}
\end{document}